# Constructing Lower Probabilities

**Carl Wagner**
Mathematics Department
University of Tennessee
Knoxville, TN 37996-1300

**Bruce Tonn**
Policy Analysis Systems Group
Oak Ridge National Laboratory
Oak Ridge, TN 37831-6207

## Abstract

An elaboration of Dempster's method of constructing belief functions suggests a broadly applicable strategy for constructing lower probabilities under a variety of evidentiary constraints.

## 1 Introduction

For any sets $X$ and $Y$ a finitely additive probability measure $p$ defined on all subsets of $Y$, together with a function $g: Y \to X$, yield a finitely additive probability measure $\pi$, defined for each $A \subset X$ by

$$\pi(A) = p(\{y \in Y : g(y) \in A\}). \tag{1}$$

This basic result of elementary probability theory furnishes a strategy for probabilistic assessment of $X$ when $X$ resists direct assessment because of its remoteness from our experience. The strategy requires identification of a set $Y$ amenable to direct assessment, as well as the determination of the function $g$, where $g(y)$ is interpreted as that outcome $x$ in $X$ implied by the outcome $y$ in $Y$.

Of course, there may in general be no such clear-cut connection between $X$ and $Y$. We might, for example, only be able to specify a function $\Gamma : Y \to 2^X - \{\emptyset\}$, with $\Gamma(y)$ being the set of outcomes in $X$ consistent with $y$. What, if any, quantification of uncertainty across $X$ can be effected in such a case?

If, paralleling (1), we define

$$\beta(A) = p(\{y \in Y : \Gamma(y) \subset A\}), \tag{2}$$

then $\beta : 2^X \to [0,1]$, $\beta(\emptyset) = 0$, and $\beta(X) = 1$, but $\beta$ is not in general additive. As first noted by Choquet (1953) in a more general setting, and as rediscovered by Dempster (1967) in the case of (2), $\beta$ is nevertheless a highly structured set function. In particular, for every integer $r \geq 2$, $\beta$ is $r$-monotone, i.e., for every sequence $A_1, \ldots, A_r$ of subsets of $X$,

$$\beta(A_1 \cup \cdots \cup A_r) \geq \sum_{\emptyset \neq I \subset [r]} (-1)^{|I|-1} \beta\left(\bigcap_{i \in I} A_i\right), \tag{3}$$

where $[r] = \{1, \ldots, r\}$. Hence $\beta$ is what Choquet calls a (normalized) *infinitely monotone capacity* and what Shafer (1976) calls a *belief function*.

Dempster calls $\beta$ the *lower probability induced by $p$ and $\Gamma$*. Although the appropriateness of this terminology is fairly clear intuitively, it is worth elaborating one precise sense in which $\beta$ may be clearly so termed. This elaboration is pursued in the next section, ultimately leading to two interesting generalizations of Dempster's construction.

## 2 An Alternative Construal of $\beta$

Although much of what follows holds in some version for arbitrary sets, we shall assume in the remainder of this paper that all sets of possible outcomes or states of affairs are *finite*. This enables us to avoid certain set theoretic complexities, such as the use of the axiom of choice. It also has the consequence that we may speak simply of probability measures, since they are identical with finitely additive probability measures on finite sets.

Suppose then that we have finite sets $X$ and $Y$, a probability measure $p$ on $2^Y$, and a consistency mapping $\Gamma : Y \to 2^X - \{\emptyset\}$, with $\beta$ defined by (2). Following Wagner (1992) we say that a probability measure $P$ on $2^{X \times Y}$ is *compatible with $p$ and $\Gamma$* if, for all $F \subset Y$,

$$P_Y(F) = p(F), \tag{4}$$

where $P_Y$ is the $Y$-marginal of $P$, i.e., $P_Y(F) = P(X \times F)$ for all $F \subset Y$, and

$$x \notin \Gamma(y) \Rightarrow P(x,y) = 0. \tag{5}$$

If we denote by $\mathcal{P}(p, \Gamma)$ the set of all probability measures $P$ on $2^{X \times Y}$ satisfying (4) and (5), then it may be shown (Wagner, 1992) that for all $A \subset X$

$$\beta(A) = \min\{P_X(A) : P \in \mathcal{P}(p, \Gamma)\}, \tag{6}$$

where $P_X(A) = P(A \times Y)$. Thus $\beta$ is simply the "lower envelope" of the family of $X$-marginals of all probability measures on $2^{X \times Y}$ compatible with $p$ and $\Gamma$.

From (6) it is clear that Dempster's construction represents just one special case of a method for assessing uncertainty across a set $X$ using a related set $Y$. The general strategy involves replacing $\mathcal{P}(p, \Gamma)$ in formula (6) with whatever family $\mathcal{P}$ of probability measures on $2^{X \times Y}$ is "compatible with the evidence," and setting

$$\lambda(A) = \min\{P_X(A) : P \in \mathcal{P}\} \quad (7)$$

for all $A \subset X$.

For what families $\mathcal{P}$ does formula (7) make sense? If $|X| = m$ and $|Y| = n$, any probability measure $P$ on $2^{X \times Y}$ is completely determined by the $mn$ values $P(x,y)$, where $x \in X$ and $y \in Y$. Any such $P$ may thus be represented as a vector in the compact subset $[0,1]^{mn}$ of $\mathbf{R}^{mn}$. Thus any family $\mathcal{P}$ of such measures corresponds to a subset of $[0,1]^{mn}$. If this subset is *closed*, and hence compact, then, for each $A \subset X$, the set $\{P_X(A) : P \in \mathcal{P}\}$ is a compact subset of $[0,1]$ (and hence possesses a minimum), since $P_X(A) = \sum_{x \in A} \sum_{y \in Y} P(x,y)$ is a continuous function of the variables $P(x,y)$.

For an arbitrary closed family $\mathcal{P}$ the determination of the minimum in (7) might of course be a difficult task. But in many applications $\mathcal{P}$ will correspond to a *closed convex polyhedral* subset of $[0,1]^{mn}$, and so the minimum can be computed by the simplex algorithm. By a closed convex polyhedral subset of $[0,1]^{mn}$ we mean, as usual, a subset defined by a finite number of linear equations and (nonstrict) inequalities on the $mn$ values $P(x,y)$. Equations and inequalities on the values of $P$ (or conditionalizations of $P$) at arbitrary subsets of $X \times Y$, or on expected values of specified random functions with respect to $P$ all fall into this category. Hence formula (7) has a wide range of practical applications.

We would emphasize that the uncertainty measure $\lambda$ defined by (7) is in general much less structured than the belief function $\beta$ arising from formulas (2) or (6). That is only to be expected, given the wide range of possible families $\mathcal{P}$. In the next section we shall examine some special families $\mathcal{P}$ for which $\lambda(A)$ may be determined by a simpler procedure than the constrained minimization required by (7). In particular, we shall derive two generalizations of Dempster's formula (2) that enable us to determine the degree of monotonicity of $\lambda$.

## 3   Some Special Cases

We begin by reviewing some basic facts about lower probabilities and lower envelopes, terms that we have been using informally, but which now require precise specification.

Let $\ell : 2^X \to [0,1]$, with $\ell(\emptyset) = 0$ and $\ell(X) = 1$, and define $u(A) := 1 - \ell(\overline{A})$ for all $A \subset X$. The set function $\ell$ is called a *lower probability* (and $u$ its corresponding *upper probability*) if, for all $A, B \subset X$, $A \cap B = \emptyset$ implies that $\ell(A \cup B) \geq \ell(A) + \ell(B)$ and $u(A \cup B) \leq u(A) + u(B)$.

If there exists a probability measure $q$ (which need not be the case) such that $q(A) \geq \ell(A)$ for all $A \subset X$ (equivalently, $q(A) \leq u(A)$ for all $A \subset X$), the lower probability $\ell$ is said to be *dominated*. A dominated lower probability $\ell$ is called a *lower envelope* if, for all $A \subset X$,

$$\begin{aligned}\ell(A) \;=\; &\min\{q(A) : q \text{ is a probability} \quad (8)\\ &\text{measure and } q(E) \geq \ell(E) \text{ for all } E \in X\}.\end{aligned}$$

If $\ell$ is a lower envelope, then for all $A \subset X$, its corresponding upper probability $u$, defined above, satisfies $u(A) = max\{q(A): q$ is a probability measure and $q(E) \leq u(E)$ for all $E \subset X\}$.

Dominated lower probabilities and lower envelopes arise naturally when $\ell(A)$ is construed behaviorally as the supremum of prices one is willing to pay to receive one unit of utility if $A$ occurs, and $u(A)$ is the infimum of payments one is willing to accept to commit oneself to pay one unit of utility if $A$ occurs. For Walley (1981) has shown that one avoids a sure loss in this context if and only if one's $\ell$ is dominated, and one avoids certain incoherent betting behavior if and only if one's $\ell$ is a lower envelope.

We now examine a special case of (7) in which the family $\mathcal{P}$ of probability measures compatible with the evidence is defined with reference to a *dominated lower probability* $\ell$ on $2^Y$, and a *family* $(\lambda_y)_{y \in Y}$ *of lower envelopes on* $2^X$, one for each $y \in Y$. We say that a probability measure $P$ on $2^{X \times Y}$ is *compatible with $\ell$ and* $(\lambda_y)_{y \in Y}$ if, for all $F \subset Y$,

$$P_Y(F) \geq \ell(F) \quad (9)$$

and also, for all $y \in Y$ with $P_Y(y) > 0$ and all $E \subset X$,

$$P(\text{``}E\text{''} \mid \text{``}y\text{''}) \geq \lambda_y(E), \quad (10)$$

where "$E$" $:= E \times Y$ and "$y$" $:= X \times \{y\}$.

We denote by $\mathcal{P}(\ell, (\lambda_y))$ the set of all $P$ satisfying (9) and (10). $\mathcal{P}(\ell, (\lambda_y))$ is always nonempty. For if we choose any probability measure $q \geq \ell$ and any family $(q_y)_{y \in Y}$ of probability measures with $q_y \geq \lambda_y$ for each $y \in Y$, and define

$$P(x,y) := q_y(x)q(y), \quad (11)$$

extending $P$ to arbitrary subsets of $X \times Y$ in the obvious way, then $P$ may easily be seen to satisfy (9) and (10).

Indeed, $\mathcal{P}(\ell, (\lambda_y))$ clearly corresponds to a closed convex polyhedral subset of $[0,1]^{mn}$, where $|X| = m$ and $|Y| = n$. Hence

$$\lambda(A) := \min\{P_X(A) : P \in \mathcal{P}(\ell, (\lambda_y))\} \quad (12)$$

may be computed by the simplex algorithm. In this case, however, the formula for $\lambda(A)$ takes a simpler form.

**Theorem 1.** *If $\lambda$ is defined by (12), then, for all $A \subset X$,*

$$\lambda(A) = \min\{\sum_{y \in Y} q(y)\lambda_y(A) : q \text{ is a} \quad (13)$$
*probability measure such that*
$$q(E) \geq \ell(E) \text{ for all } E \subset X\}.$$

Proof. Note first that the right hand side of (13) is well defined, since $\sum_{y \in Y} q(y)\lambda_y(A)$ is a continuous function of the $n$ variables $q(y)$, with domain a closed convex polyhedral subset of $[0,1]^n$. Hence establishing (13), with its reduced number of variables and constraints, will considerably simplify the calculation of $\lambda(A)$.

To prove (13), define the subsets $U_A$ and $V_A$ of $[0,1]$ by
$$U_A = \{P_X(A) : P \in \mathcal{P}(\ell, (\lambda_y))\} \quad (14)$$
and
$$V_A = \{\sum_{y \in Y} q(y)\lambda_y(A) : \quad (15)$$
$$\text{probability measure } q \geq \ell.\}$$

We must show that $\min U_A = \min V_A$. We do this by proving that (i) for every $u \in U_A$, there exists a $v \in V_A$ such that $v \leq u$ (whence $\min U_A \geq \min V_A$) and that (ii) $V_A \subset U_A$ (whence $\min U_A \leq \min V_A$).

To prove (i), it suffices to show that if $P \in \mathcal{P}(\ell, (\lambda_y))$, then
$$P_X(A) \geq \sum_{y \in Y} P_Y(y)\lambda_y(A), \quad (16)$$
since by hypothesis $P_Y \geq \ell$. Writing "$A$" for $A \times Y$ and "$y$" for $X \times \{y\}$, we have

$$P_X(A) = P(\text{``}A\text{''}) = \sum_{y \in Y : P(\text{``}y\text{''}) > 0} P(\text{``}A\text{''} \cap \text{``}y\text{''})$$
$$= \sum_{y \in Y : P(\text{``}y\text{''}) > 0} P(\text{``}y\text{''}) P(\text{``}A\text{''} \mid \text{``}y\text{''})$$
$$\geq \sum_{y \in Y} P(\text{``}y\text{''})\lambda_y(A) = \sum_{y \in Y} P_Y(y)\lambda_y(A),$$

as desired.

To prove (ii), we must show that for every $q \geq \ell$, there exists a $P \in \mathcal{P}(\ell, (\lambda_y))$ such that
$$P_X(A) = \sum_{y \in Y} q(y)\lambda_y(A). \quad (17)$$
To construct such a $P$ choose a family of probability measures $(q_y)_{y \in Y}$ on $2^X$ such that $q_y \geq \lambda_y$ and $q_y(A) = \lambda_y(A)$, this being possible since each $\lambda_y$ is a lower envelope. We have noted above that setting $P(x, y) = q_y(x)q(y)$ yields a $P \in \mathcal{P}(\ell, (\lambda_y))$. It is also the case that
$$P_X(A) = \sum_{y \in Y}\sum_{x \in A} P(x, y)$$
$$= \sum_{y \in Y} q(y) \sum_{x \in A} q_y(x)$$
$$= \sum_{y \in Y} q(y)q_y(A) = \sum_{y \in Y} q(y)\lambda_y(A),$$

which completes the proof.

We conclude by examining two special cases of formula (13) for which there are formulas for $\lambda(A)$ not involving any minimization operation. The two formulas are each generalizations of Dempster's formula (2).

**(I).** In the first case the dominated lower probability $\ell$ is actually a probability measure, call it $p$. Since the only probability measure $q \geq p$ is $p$ itself, $V_A$, as defined by (15), has a single element, $\sum_{y \in Y} p(y)\lambda_y(A)$, and so formula (13) becomes
$$\lambda(A) = \sum_{y \in Y} p(y)\lambda_y(A). \quad (18)$$

It is easy to see that $\lambda$ is $r$-monotone if each $\lambda_y$ is $r$-monotone. In particular, $\lambda$ is a belief function if each $\lambda_y$ is a belief function.

To see how (18) reduces to (2), consider the special family $(\lambda_y)_{y \in Y}$ defined in terms of a family $(E_y)_{y \in Y}$ of nonempty subsets of $X$ by
$$\lambda_y(A) = \begin{cases} 1, & \text{if } A \supset E_y \\ 0, & \text{otherwise.} \end{cases} \quad (19)$$

These functions $\lambda_y$, called *simple support functions* by Shafer, are belief functions, and clearly lower envelopes. Define $\Gamma : Y \to 2^X - \{\emptyset\}$ by $\Gamma(y) = E_y$ for all $y \in Y$. Then (18) becomes
$$\lambda(A) = \sum_{y \in Y} p(y)\lambda_y(A) = \sum_{y \in Y : \Gamma(y) \subset A} p(y)$$
$$= p(\{y \in Y : \Gamma(y) \subset A\}),$$
which is (2).

**(II).** In the second special case of (13), we take for $\ell$ any lower envelope and let the family $(\lambda_y)_{y \in Y}$ be defined by (19), with $\Gamma$ defined as above. Then formula (13) becomes
$$\lambda(A) = \ell(\{y \in Y : \Gamma(y) \subset A\}), \quad (20)$$
by the following derivation:

$$\lambda(A) = \min\{\sum_{y \in Y} q(y)\lambda_y(A) : \text{ probability } q \geq \ell\}$$
$$= \min\left\{\sum_{y \in Y : \Gamma(y) \subset A} q(y) : \text{ probability } q \geq \ell\right\}$$
$$= \min\{q(\{y \in Y : \Gamma(y) \subset A\}) :$$
$$\text{probability } q \geq \ell\}$$
$$= \ell(\{y \in Y : \Gamma(y) \subset A\}),$$

the last equality following from the fact that $\ell$ is a lower envelope. Of course (20) reduces to Dempster's formula (2) when $\ell$ is actually a probability measure $p$.

We leave it as an exercise to show that if $\ell$ is $r$-monotone, then $\lambda$, as given by (20), is also $r$-monotone. In particular, if $\ell$ is a belief function, then so is $\lambda$, as Shafer (1979) has previously observed.

## 4    Acknowledgements

Wagner's research was supported in part by the National Science Foundation (DIR-8921269).